\documentclass[10pt,twocolumn,letterpaper]{ieeeconf}

\usepackage{graphicx}
\usepackage{subcaption}
\usepackage{multicol}
\usepackage{multirow}
\usepackage{mathtools}
\usepackage{algorithm}
\usepackage{algorithmic}
\usepackage{pifont}% Zapf-Dingbats Symbole
\newcommand*{\FeatureTrue}{\ding{52}}
\newcommand*{\FeatureFalse}{\ding{56}}
\PassOptionsToPackage{hyphens}{url}\usepackage[pagebackref=true,breaklinks=true,letterpaper=true,colorlinks,bookmarks=false]{hyperref}

\title{\LARGE \bf
Unsupervised Face Recognition using Unlabeled Synthetic Data 
}
\author{\parbox{16cm}{\centering
    {\large Fadi Boutros$^{1,2}$, Marcel Klemt$^1$, Meiling Fang$^{1,2}$, Arjan Kuijper$^{1,2}$ and Naser Damer$^{1,2}$}\\
    {\normalsize
    $^1$Fraunhofer Institute for Computer Graphics Research IGD,
Darmstadt, Germany\\
$^{2}$Department of Computer Science, TU Darmstadt,
Darmstadt, Germany\\
 Email: {fadi.boutros@igd.fraunhofer.de}
 }}
    \thanks{This research work has been funded by the German Federal Ministry of Education and Research and the Hessen State Ministry for Higher Education, Research and the Arts within their joint support of the National Research Center for Applied Cybersecurity ATHENE. This work has been partially funded by the German Federal Ministry of Education and Research (BMBF) through the Software Campus Project.}% <-this % stops a space
}

\begin{document}

%\IEEEoverridecommandlockouts\pubid{\makebox[\columnwidth]{979-8-3503-4544-5/23/\$31.00~\copyright{}2023 IEEE \hfill}
%\hspace{\columnsep}\makebox[\columnwidth]{ }}

%\ifFGfinal
%\thispagestyle{empty}
%\pagestyle{empty}
%\else
%\author{Anonymous FG2023 submission\\ Paper ID \FGPaperID \\}
%\pagestyle{plain}
%\fi
\maketitle

%%%%%%%%%%%%%%%%%%%%%%%%%%%%%%%%%%%%%%%%%%%%%%%%%%%%%%%%%%%%%%%%%%%%%%%%%%%%%%%%
\begin{abstract}
Over the past years, the main research innovations in face recognition focused on training deep neural networks on large-scale identity-labeled datasets using variations of multi-class classification losses. However, many of these datasets are retreated by their creators due to increased privacy and ethical concerns. Very recently, privacy-friendly synthetic data has been proposed as an alternative to privacy-sensitive authentic data to comply with privacy regulations and to ensure the continuity of face recognition research. In this paper, we propose an unsupervised face recognition model based on unlabeled synthetic data (USynthFace). 
Our proposed USynthFace learns to maximize the similarity between two augmented images of the same synthetic instance.  
We enable this by a large set of geometric and color transformations in addition to GAN-based augmentation that contributes to the USynthFace model training. 
We also conduct numerous empirical studies on different components of our USynthFace. With the proposed set of augmentation operations, we proved the effectiveness of our USynthFace in achieving relatively high recognition accuracies using unlabeled synthetic data. The training code and pretrained model are publicly available under \url{https://github.com/fdbtrs/Unsupervised-Face-Recognition-using-Unlabeled-Synthetic-Data}.
%Through extensive evaluation experiments on five main benchmarks, we 

\end{abstract}

%%%%%%%%%%%%%%%%%%%%%%%%%%%%%%%%%%%%%%%%%%%%%%%%%%%%%%%%%%%%%%%%%%%%%%%%%%%%%%%%

\section{Introduction}
% 1.  why do we need synthetic training? Motivation + data availability + data labeled
The evolution in deep learning network architectures, training losses, and availability of large-scale identity-labeled training datasets are behind the major advances in recognition accuracy by the recent state-of-the-art (SOTA) face recognition (FR) models.  
The main FR works focus on proposing novel FR training losses, especially margin-penalty based softmax loss e.g. ArcFace \cite{ArcFace}, CurricularFace \cite{CurricularFace} or ElasticFace \cite{ElasticFace}, to train deep neural network \cite{ResNet} on large-scale identity-labeled dataset \cite{CASIA, MS-Celeb-1M, VGGFace2}. This research direction is driven by the availability of large-scale identity-labeled datasets and the high recognition performance achieved by margin-penalty softmax losses. Recently, there were an increase concerns about collecting, maintaining, redistributing and using biometric data due to legal and ethical privacy issues in some countries \cite{DBLP:journals/tifs/MedenRTDKSRPS21,eu-2016/679}. Especially that many of FR datasets such as VGGFace2 \cite{VGGFace2} have been collected from the web without the proper consent of subjects.
Privacy regulations such as the General Data Protection Regulation (GDPR) \cite{eu-2016/679} classify biometric data as personal data. They grant the right to individuals to withdraw their consent to use or store their personal data. Practically, maintaining such regulations is challenging, especially in the case that such data is collected from the web and is widely distributed.  

To overcome this challenge, the use of privacy-friendly synthetic data as an alternative to authentic data in biometrics development has recently attracted attention \cite{DBLP:journals/ivc/BoutrosDRRKK20,quantface_boutros,DBLP:conf/cvpr/DamerLFSPB22,DBLP:conf/iccvw/DamerBKK19}.
In the field of FR, two main previous works proposed the use of synthetically generated data by Generative Adversarial Network (GAN) \cite{GAN} to develop FR models. SynFace \cite{SynFace} investigated the different behavior of FR models trained on authentic and synthetic images and proposed identity and domain mixup to reduce the performance gap of FR models trained on synthetic data in comparison to FR models trained on authentic data. SFace \cite{SFace} proposed the use of synthetic data to develop FR models, including presenting a public synthetic database, FR training protocols, detailed analyses of the identity transfer from generator training to the generated data, the identifiability of the authentic data in the trained models. SynFace \cite{SynFace} and SFace \cite{SFace} mainly focused on using synthetic data to train supervised FR to learn multi-class classification problems. %However, such solution are limited to  

\begin{figure*}[t!]
	\centering
	\includegraphics[width=0.85\linewidth]{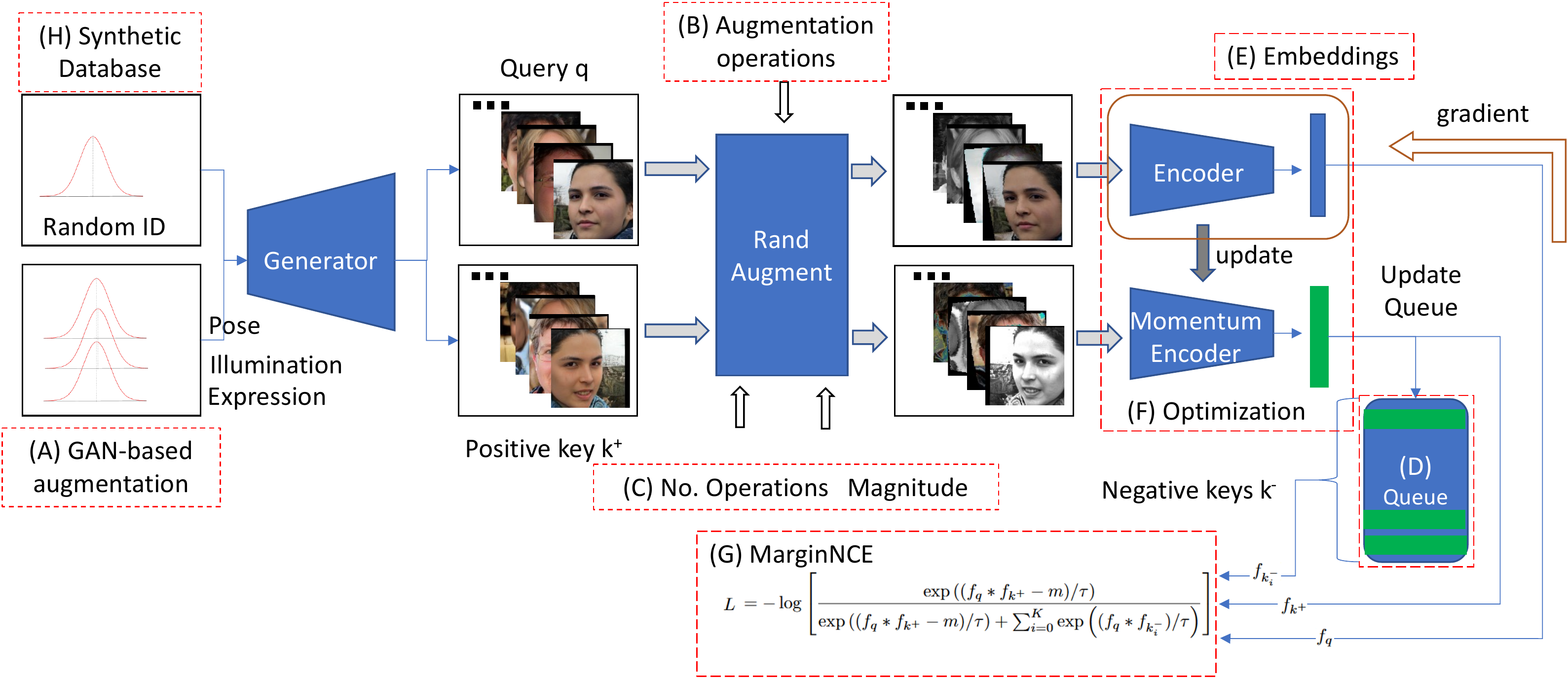}
	\vspace{2mm}
    \caption{An overview of our unsupervised FR training paradigm, USynthFace. Synthetic samples are generated and augmented by a generative model to output a query ($q$) and a positive key $k^+$. Then, these samples are augmented with color and geometric transformation through RandAugment. The augmented $q$ and $k^+$ are processed by the encoder and the momentum encoder, respectively. Then, the queue is updated where the current mini-batch is enqueued and the oldest mini-batch from the queue is dequeued. Finally, 
    the contrastive loss is calculated based on $q$, $k^+$ and the negative keys ($k^-$). Gradients are only back-propagated through the encoder. %(i.e. our USynthFace FR model). 
    To maintain key consistency in the queue and to avoid rapidly updating the key encoder, updating the key encoder (momentum encoder) is driven by momentum update with the query encoder.
    %the key encoder is slowly updated with momentum
    %The momentum encoder is slowly updated
    %with slowley
    %by a fraction of the encoder new parameters'.
    }
	\label{fig:overview}
	%\vspace{-5mm}
\end{figure*}

This work presents contributions toward developing unsupervised FR using privacy-friendly synthetic data (USynthFace).  
Unlike previous works that require synthetic labeled data \cite{SFace} or mixing up authentic with synthetic data \cite{SynFace}, our proposed framework does not require labeled data or involve authentic data in the FR model training. Thus, it takes full advantage of privacy-friendly synthetic data and does not require a special GAN model to generate labeled data. 
This work is the first to propose unsupervised FR using synthetic data. Our unsupervised FR training paradigm is based on the concept of the Momentum Contrast \cite{Moco} and contrastive learning \cite{InfoNCE} for unsupervised representation learning. 
The main idea of our approach is to extract a pair of feature representations from two augmented views of the same instance. Then,  learning to enhance the similarity between this pair to be higher than the similarity to any other instance. As such learning paradigm mainly depends on augmenting the training sample, we propose a large set of augmentation operations based on geometric and color transformations, as well as controlled GAN augmentation to simulate different realistic appearance variations, i.e. pose, illumination, and expression. We also provide sensitivity studies on all the components of our unsupervised FR framework.  USynthFace achieved relatively high verification performances on several benchmarks. Our USynthFace also achieved very competitive results to supervised FR trained on synthetic data and outperformed them on several benchmarks.  For example, our USynthFace achieved SOTA accuracy (92.23\%) on Labeled Face on the Wild (LFW) for FR trained on synthetic data.

\section{Methodology}
This section presents our unsupervised FR training framework and its components based on synthetic data (USynthFace). Figure \ref{fig:overview} illustrates the pipeline of our proposed framework. Synthetic face images are conditionally generated using conditional GAN with different random identities, pose, illumination and expression to output a query ($q$) and a positive key ($k^+$), which are two instances sampled using same random identity latent vector with different pose, illumination and expression. Then, the query and the positive key are further augmented with geometric and color transformations. The query images are then processed by the encoder and the positive keys are processed by the momentum encoder. The resulting feature representations of the momentum encoder are then pushed into the queue and the ones from the oldest batch are dequeued. Finally, the feature representations of the query, the positive key and the negative keys (retrieved from the queue) are utilized to calculate the contrastive loss.

\subsection{Unsupervised face recognition}

\paragraph{Unsupervised representation learning}

%shortly explain what is moco
%how it is different than other unsupervised approaches

We utilize in this work the concept of the Momentum Contrast (MoCo)\cite{Moco} for unsupervised representation learning.
MoCo uses contrastive learning to maximize the similarity between feature representations of positive pairs and minimize the similarity between feature representations of negative pairs. Positive pairs are two versions of the same instance augmented using geometric or color transformations, while pairing with any other images (of different instances) is considered as negative pair, which might be perceived as self-supervised learning. MoCo introduced a dynamic queue to form a larger amount of negative pairs than forming negative pairs only from the current batch \cite{DBLP:conf/cvpr/YeZYC19,DBLP:conf/iclr/HjelmFLGBTB19}. Moreover, unlike memory bank where the feature encodings are produced from different training stages \cite{DBLP:conf/cvpr/WuXYL18}, Moco introduced momentum encoder to maintain consistent feature representations.

Consider a query image $q$ encoded into $f_q$, a positive key of the same instance of $q$, augmented as $k^+$ and encoded into $f_{k^+}$ along with a set of negative keys $\{k^-_i\}^K_{i=1}$ (encoded into $\{f_{k^-_i}\}^K_{i=1}$) that are retrieved from the queue. A contrastive loss that guides the model to enhance the similarity between $f_q$ and $f_{k^+}$ to be larger than the similarity between $f_q$ and $\{f_{k^-_i}\}^K_{i=1}$ can be measured using MarginNCE \cite{MarginNCE} as follows:
\begin{footnotesize}
\begin{align}
    L = - \log\frac{\exp\left((f_q*f_{k^+} - m) / \tau \right)} {\exp\left((f_q*f_{k^+} - m) / \tau \right) + \sum_{i=1}^{K}\exp\left((f_q*f_{k^-_i} ) / \tau\right)}
    \label{eq:MarginNCE}
\end{align}
\end{footnotesize}
where $\tau$ is a temperature hyper-parameter that controls the entropy of the distribution \cite{KD_temperature} and $m$ is a margin penalty used to encourage the model to learn discriminative feature representations. A detailed sensitivity study on the optimal margin selection is provided in Section \ref{sec:abl_margin}.
Once the loss function is calculated, the loss gradients are only back-propagated through the encoder (query encoder $\theta_{enc}$). 
To avoid rapidly updating the key encoder which might break the queue consistency \cite{Moco}, the weights of the momentum encoder (key encoder $\theta_{mom\_enc}$) are slowly updated by evolving the query encoder \cite{Moco} with a momentum coefficient, as follows: $\theta_{mom\_enc} \leftarrow mc * \theta_{mom\_enc} + (1-mc)* \theta_{enc}$, where $mc \in [0,1]$ is a momentum coefficient. %We followed \cite{Moco, MocoV2, UniMoCo} to set $mc$ to $0.999$ in all our experiments.

\begin{algorithm}
\caption{USynthFace training pipeline}
\begin{algorithmic} 
\small
\STATE $Z_{id} \leftarrow$ sample $I$ vectors from $N(0,1)$
\STATE $RA \leftarrow RandAugment(N,M)$
\WHILE{$e < num\_epochs$}
\STATE shuffle $Z_{id}$
\FORALL{$z_{id}$ in $Z_{id}$}
\FOR{$i$ in $[0,1]$}
\STATE $z_{pose}, z_{expr}, z_{illu} \sim N(0,1)$
\STATE $z_{app}^{(i)} \leftarrow z_{pose} \mathbin\Vert z_{expr} \mathbin\Vert z_{illu}$
\ENDFOR
\STATE $q \leftarrow G(z_{id} \mathbin\Vert z_{app}^{(0)})$
\STATE $k^+ \leftarrow G(z_{id} \mathbin\Vert z_{app}^{(1)})$
\STATE $q \leftarrow RA(q)$
\STATE $k^+ \leftarrow RA(k^+)$
\STATE $f_q \leftarrow enc(q)$
\STATE $\theta_{mom\_enc} \leftarrow mc * \theta_{mom\_enc} + (1-mc)* \theta_{enc}$
\STATE $f_{k^+} \leftarrow mom\_enc(k^+)$
\STATE $queue \leftarrow update(f_{k^+}, queue)$
\STATE $f_{k_i^-} \leftarrow queue$
\STATE $l \leftarrow - \log\frac{\exp\left((f_q*f_{k^+} - m) / \tau \right)} {\exp\left((f_q*f_{k^+} - m) / \tau \right) + \sum_{i=1}^{K}\exp\left((f_q*f_{k^-_i} ) / \tau\right)}$
\STATE $backward(enc, l)$
\ENDFOR
\ENDWHILE
\end{algorithmic}
% \caption{Pseudocode for training FaceMoCo. $G$ describes the generator which generates image from the latent representation. $enc$ and $mom\_enc$ stands for encoder and momentum encoder and $mc$ is the momentum coefficient by which the momentum encoder is updated.}
\end{algorithm}
%\vspace{-4mm}

\begin{figure}[ht!]
	\centering
	\includegraphics[width=0.70\linewidth]{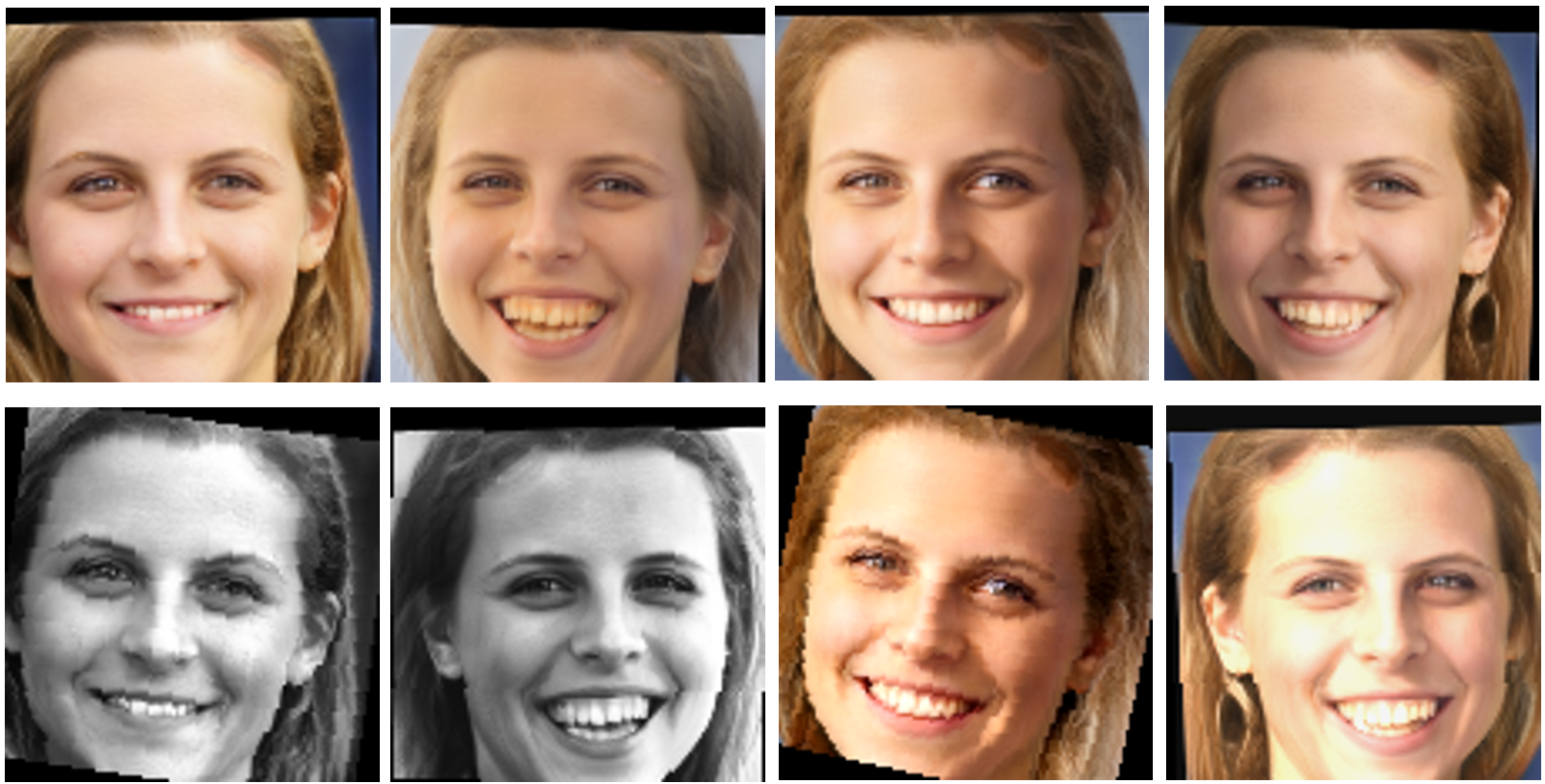}
    \caption{Samples of augmented images.
    %GAN-based (first row) and RandAugment (second row) augmented images. 
    The first row shows synthetically generated images with a same identity latent vector and the augmented version of these images with RandAugment are presented in the second row. %The second row shows the exact images from  augmented via RandAugment.
    }
	\label{fig:GAN_and_RA_augmentations}
	\vspace{-4mm}
\end{figure}

\paragraph{Synthetic data generation}
\label{sec:syn_data_generation}
The training dataset of our unsupervised model is synthetically generated using Generative Adversarial Network (GAN) \cite{GAN}. Specifically, we used DiscoFaceGAN (DFG) \cite{DiscoFaceGAN} to conditionally generate $I$ images with different identity, pose, illumination, and expression. 
DFG presented a 3D morphable face model (3DMM) \cite{3DMM} to the StyleGAN model \cite{StyleGAN}, enabling disentanglement of identity, pose, expression and illumination in the latent space to conditionally generate realistic images with varying attributes. 
The conditional image generation serves as data augmentation for our unsupervised learning model as presented in the next section. 

%These images are synthetically augmented by fixing the identity latent representation and changing the pose, illumination and expression as detailed in the next section.    

%how did we genererate the data... discoFacegan based on styleGan2

%Our GAN-based augmented images are generated using DiscoFaceGAN (DFG) \cite{DiscoFaceGAN}. DFG includes a 3D morphable face model (3DMM) \cite{3DMM} to their StyleGAN-based generator model \cite{StyleGAN} to enforce a disentanglement of the latent space. This allows a generation of images of the same class with varying pose, expression, and illumination.

\paragraph{Data augmentation}
\label{sec:data_augmentation}
Unsupervised representation learning approaches such as MoCo \cite{Moco}, AMDIM \cite{DBLP:conf/nips/BachmanHB19}, and SimCLR \cite{DBLP:conf/icml/ChenK0H20} are heavily dependant on data augmentation to construct positive pairs of the same instance. Previous approaches \cite{DBLP:conf/icml/ChenK0H20,DBLP:conf/nips/BachmanHB19,Moco} utilized geometric and color transformation for augmenting the training data. 
In this work, we propose to enrich the conventional data augmentation operations, i.e. geometric and color transformations with GAN-based augmentations generated by a conditional generative model.
%based pose, illumination and expression to simulate real  
The conventional data augmentation method is based on RandAugment \cite{Randaugment}. The search space of RandAugment has two hyperparameters $N$ and $M$, where $N$ is the number of transformations applied sequentially to each input image and $M$ is the magnitude of each transformation. 
%In contrast to default RandAugment, we randomly sample the applied magnitude from a range of [1, $M$] instead of applying always the same $M$ value. This allows the model the generalize to a wider variety of views.
Transformation operations include: Horizontal-flipping, Rotate, Translate-x, Translate-y, Shear-x, Shear-y, Sharpness, AutoContrast, Contrast, Solarize, Posterize, Equalize, Color, Brightness, ResizedCrop and Grayscale. Sensitive studies on the effect of each operation on FR verification performance and the selection of RandAugment optimal hyper-parameters are provided in Sections \ref{sec:abl_RA_operations} and \ref{sec:abl_RA_N_M}, respectively.
To simulate more variations that occur in real images, we also utilize DFG \cite{DiscoFaceGAN} to augment training images with different pose, facial expression, and illumination. 
To generate such images, we first randomly sample 128-D vector from a normal Gaussian distribution $N(0,1)$, which represents the identity information \cite{DiscoFaceGAN}. 
The expression, illumination and pose attributes are controlled by three latent vectors (32-D, 16-D and 3-D, respectively) and are disentangled from the identity latent vector. Two augmented views of the same image (and thus identity) can be generated by fixing the identity latent and randomly modifying the attribute latent vectors. Formally, two augmented images, query $q$ and positive key $k^+$, can be generated by sampling two appearance vectors as follows:
\begin{equation}
    z_{app}^{(0)} = z_{pose}\mathbin\Vert z_{expr}\mathbin\Vert z_{ill}, \{z_{pose}, z_{expr},  z_{ill}\} \sim N(0,1)
\end{equation}
    and
\begin{equation}
    z_{app}^{(1)} = z_{pose}\mathbin\Vert z_{expr}\mathbin\Vert z_{ill}, \{z_{pose}, z_{expr},  z_{ill}\} \sim N(0,1).
\end{equation}
Each of $z_{app}^{(0)}$ and $z_{app}^{(1)}$ are then concatenated with identity latent vector $z_{id} \sim N(0,1)$ (randomly sampled) to generate augmented $q$ and $k^+$, as follows:
\begin{equation}
   q = G(z_{id}\mathbin\Vert z_{app}^{(0)})
\end{equation}
   and
\begin{equation}
   k^+ = G(z_{id}\mathbin\Vert z_{app}^{(1)}).
\end{equation}

\begin{figure*}[ht!]
	\centering
    \begin{subfigure}[b]{0.24\textwidth}
        \centering
        \includegraphics[width=\textwidth]{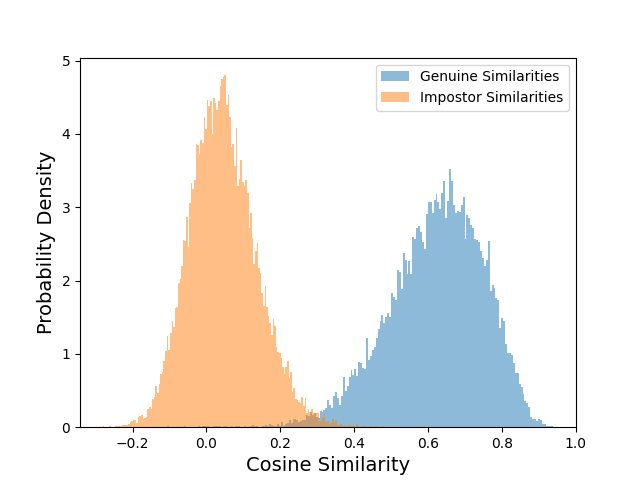}
        \caption{GAN-based.}
        \label{fig:DFG_distribution}
    \end{subfigure}
    %\hfill
    \begin{subfigure}[b]{0.24\textwidth}
        \centering
        \includegraphics[width=\textwidth]{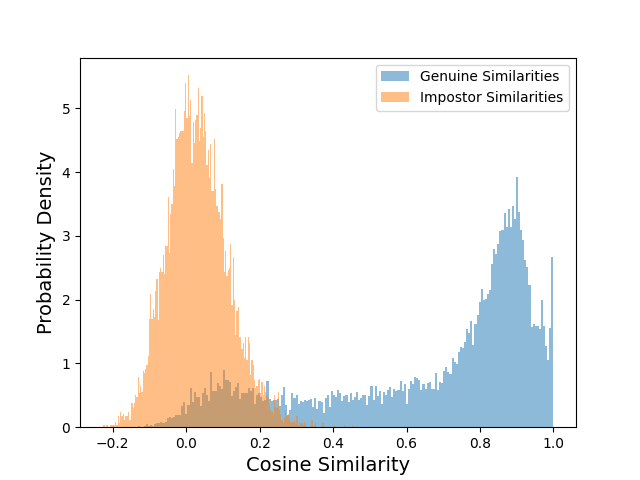}
        \caption{RandAugment}
        \label{fig:RA_distribution}
    \end{subfigure}
    %\hfill
    \begin{subfigure}[b]{0.24\textwidth}
        \centering
        \includegraphics[width=\textwidth]{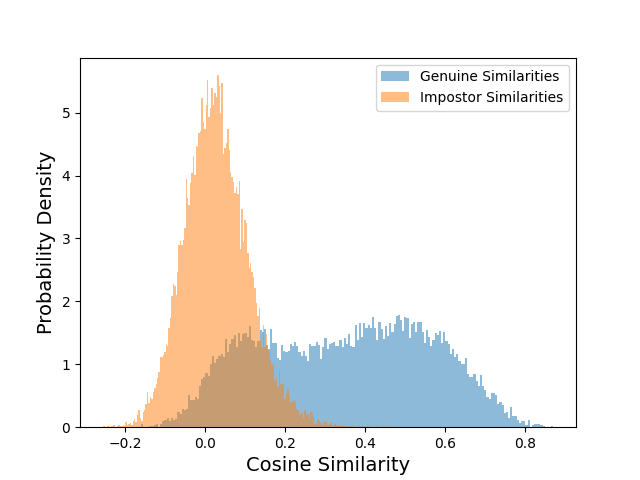}
        \caption{GAN-based + RandAugment}
        \label{fig:DFG_RA_distribution}
    \end{subfigure}
    \vspace{-2mm}
	\caption{The genuine (blue) and imposter (orange) score distributions of three different data augmentation settings. The genuine pairs are constructed using GAN-based augmentation (Figure \ref{fig:DFG_distribution}), RandAugment (Figure \ref{fig:RA_distribution}), and GAN-based with RandAugment (Figure \ref{fig:DFG_RA_distribution}). The biggest effect is noticed when combining both augmentations.}
	\label{fig:identity_distributions}
	\vspace{-4mm}
\end{figure*}

\begin{table}[h!]
\resizebox{\linewidth}{!}{%
\begin{tabular}{|c|c|c|c|c|}
\hline
\textbf{Augmentation}   & \textbf{EER} & \textbf{FMR10} & \textbf{FMR100} & \textbf{FMR1000} \\ \hline
GAN-based               & 0.0110       & 0.0019         & 0.0118          & 0.0558           \\ \hline
RandAugment             & 0.0967       & 0.0955         & 0.1520          & 0.1915           \\ \hline
GAN-based + RandAugment & 0.1650       & 0.2038         & 0.3547          & 0.4681           \\ \hline
\end{tabular}%
}
\caption{The effect of augmentation on identity in the images indicated by the verification performances as EER, FMR10, FMR100 and FMR1000 on three constructed datasets using GAN-based, RandAugment, and GAN-based with RandAugment augmentations. GAN-based with RandAugment augmentation results in bigger effects on identity, providing more challenging samples for the FR training.}
\label{tab:distribution_metrics}
\vspace{-4mm}
\end{table}

\section{Experimental setups}
This section presents the experimental settings followed in this paper.
\subsection{Dataset}
\vspace{-2mm}
\paragraph{Training dataset}
We employ a pretrained DFG model to synthetically generate facial images as discussed in Section \ref{sec:syn_data_generation}. The model is trained on Flickr-Faces-HQ dataset (FFHQ) \cite{StyleGAN} that contains 70k images of the size $1024 \times 1024$ pixels collected from Flickr and encompass variation in ethnicity, age, image background, and accessories \cite{StyleGAN}. We opt to generate 100K images from the DFG model, each from different identity latent representations. During the training phase, we augmented these images with conventional augmentation transformations as well as with GAN-based augmentation i.e. pose, illumination and expression (as detailed in Section \ref{sec:data_augmentation}). All training data are aligned and cropped to $112 \times 112$ using similarity transformation \cite{ArcFace,ElasticFace,CurricularFace} based on detected facial landmarks by Multi-task Cascaded Convolutional Networks (MTCNN) \cite{MTCNN}. All images are then normalized to have pixel values between -1 and 1.

\paragraph{Evaluation datasets}
We used in this paper the following datasets as evaluation benchmarks for our ablation studies: Labeled Faces in the Wild (LFW) \cite{LFW}, AgeDB-30 \cite{agedb}, Celebrities in Frontal to Profile in the Wild (CFP-FP) \cite{cfpfp}, Cross-Age LFW (CA-LFW) \cite{calfw}, and Cross-Pose LFW (CP-LFW) \cite{cplfw}. The verification accuracy is reported for each of the considered benchmarks following their defined protocols. In all ablation studies in this paper, the overall verification performance is based on the sum of the performance ranking Borda count on LFW, AgeDB-30, CFP-FP, CA-LFW and CP-LFW.
% The final model is additionally evaluated on the large-scale IARPA Janus Benchmark-B (IJB-B) \cite{ijbb} and Benchmark-C (IJB-C) \cite{ijbc}. Following the official 1:N verification protocols of IJB-B and IJB-C, the verification performance is reported as true acceptance rates (TAR) at false acceptance rates (FAR) of 1e-4, 1e-5 and 1e-6.
%are utilized as validation datasets. Following their protocol, verification accuracy is reported.

%which was originally created for GAN training consisting of 70,000 face images without identity labels. The images were collected from Flickr and encompass variation in ethnicity, age, image background, and accessories. All images were aligned and cropped using the dlib library to 1024x1024 pixel.

\begin{table}[ht]
\resizebox{\linewidth}{!}{%
\begin{tabular}{|c|cc|cc|}
\hline
\multirow{2}{*}{\textbf{Model}} & \multicolumn{2}{c|}{\textbf{R-R} (\%)}                                                           & \multicolumn{2}{c|}{\textbf{R-S} (\%)}                                                           \\ \cline{2-5} 
                                & \multicolumn{1}{c|}{\textbf{\textgreater{}FMR100\_Th}} & \textbf{\textgreater{}FMR1000\_Th} & \multicolumn{1}{c|}{\textbf{\textgreater{}FMR100\_Th}} & \textbf{\textgreater{}FMR1000\_Th} \\ \hline
ArcFace                         & \multicolumn{1}{c|}{2.6857}                            & 0.5664                             & \multicolumn{1}{c|}{3.1015}                            & 0.5827                             \\ \hline
CurricularFace                  & \multicolumn{1}{c|}{1.9137}                            & 0.4024                             & \multicolumn{1}{c|}{2.0284}                            & 0.3741                             \\ \hline
ElasticFace                     & \multicolumn{1}{c|}{2.0538}                            & 0.2951                             & \multicolumn{1}{c|}{2.3518}                            & 0.3130                             \\ \hline
\end{tabular}%
}
\caption{Percentage of comparison scores that are larger than different operation thresholds at FMR100 and FMR1000 for R-R and R-S impostor comparison. The percentage number indicates how many comparisons are falsely matched as genuine. The low percentage for the R-S setting and its similarity to R-R indicates that the identities of the authentic data is not linked to these of the synthetic data.}
\label{tab:percentage_above_th}
\vspace{-4mm}
\end{table}

\subsection{Model training setup}
The network architecture of the encoder model is ResNet-50 \cite{ResNet}, which is one of the widely used architectures in recent SOTA FR \cite{ArcFace, PartialFC, CurricularFace, ElasticFace}. Following \cite{Moco}, the momentum encoder is updated with a momentum coefficient of 0.999 \cite{Moco} and the temperature value $\tau$ of contrastive loss is set to 0.07 \cite{Moco}. The feature representation dimensions is initially set to 512-D in the results presented in Sections \ref{sec:abl_RA_operations}, \ref{sec:abl_RA_N_M}, \ref{sec:abl_queue_size}.
Later, in Section \ref{sec:abl_feat_dim} we present an ablation study on the optimal feature representation dimensions of 128, 256, 512, and 1024-D. The queue size is set to 32768 based on sensitivity study presented in Section \ref{sec:abl_queue_size}. 
%As encoder and momentum encoder, ResNet-50 \cite{ResNet} is employed as this architecture is utilized in MoCo \cite{Moco} and widely used in SOTA FR \cite{ArcFace, PartialFC, CurricularFace, ElasticFace}. 
%The momentum encoder is updated with a momentum coefficient of 0.999 as proposed in MoCo. InfoNCE loss \cite{InfoNCE} is also adopted with a temperature value $\tau$ of 0.07 and the queue size is set to 65,536. The feature dimension is increased to 512 as it is common in FR \cite{SphereFace, CosFace, ArcFace, ElasticFace}. 
An optimizer Stochastic Gradient Descent (SGD) is used with initial learning rate of 0.1. The momentum is set to 0.9 and the weight decay to 5e-4. The learning rate is divided by 10 after 8, 16, 24, and 32 epochs. The models presented in Sections \ref{sec:abl_RA_operations}, \ref{sec:abl_RA_N_M}, \ref{sec:abl_queue_size}, \ref{sec:abl_feat_dim} are trained for 40 epochs in total with a batch size of 512 on 100K synthetic images. 
All models are implemented using PyTorch \cite{PyTorch} and trained on two CPU 16 core Intel Xeon Gold 5218 and four NVIDIA Quadro RTX6000 GPUs.

\begin{figure*}[ht!]
	\centering
    \begin{subfigure}[b]{0.25\linewidth}
        \centering
        \includegraphics[width=\textwidth]{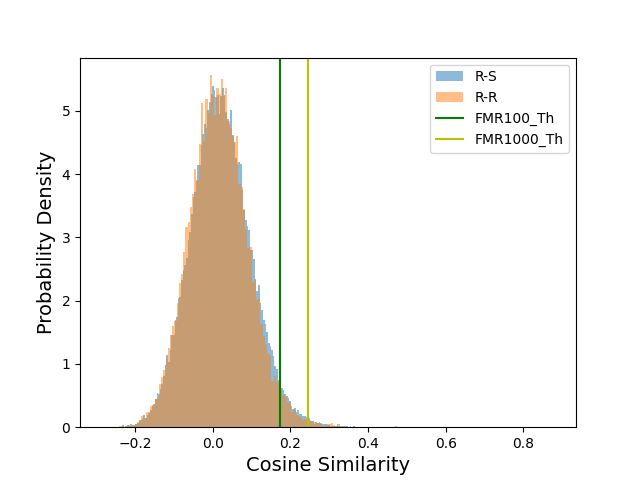}
        \caption{ArcFace}
        \label{fig:R-S_ArcFace}
    \end{subfigure}
    %\hfill
    \begin{subfigure}[b]{0.25\linewidth}
        \centering
        \includegraphics[width=\textwidth]{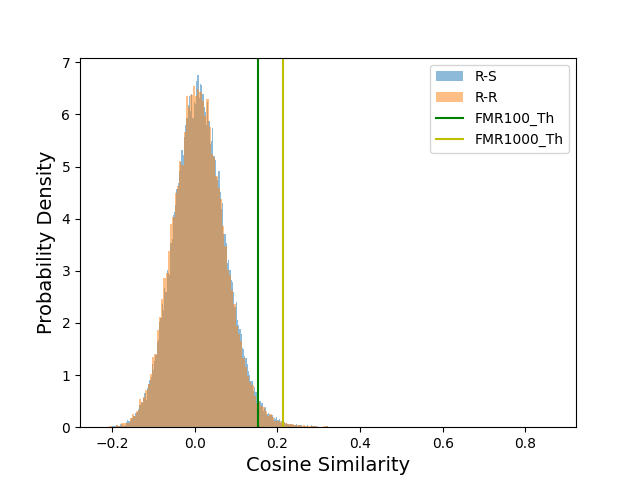}
        \caption{CurricularFace}
        \label{fig:R-S_CurricularFace}
    \end{subfigure}
    %\hfill
    \begin{subfigure}[b]{0.25\linewidth}
        \centering
        \includegraphics[width=\textwidth]{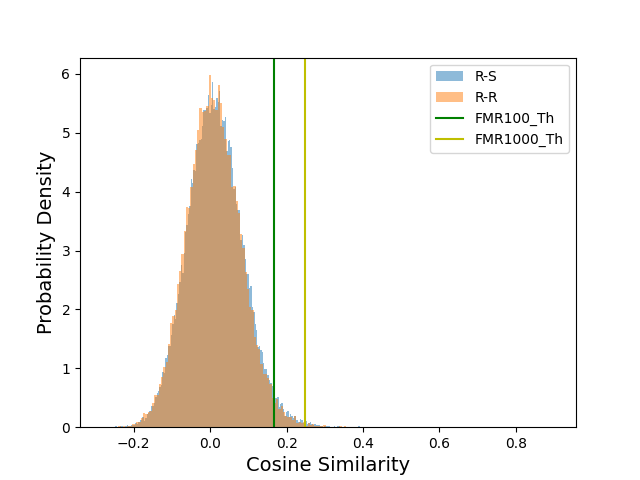}
        \caption{ElasticFace}
        \label{fig:R-S_ElasticFace}
    \end{subfigure}
    \vspace{-3mm}
	\caption{The score distributions of the two settings, the authentic data R-R and the cross-dataset R-S, achieved by ArcFace \cite{ArcFace}, CurricularFace \cite{CurricularFace}, and ElasticFace \cite{ElasticFace}. The highly overlapping score distributions indicate an extremely weak identity relation between the authentic training data and the generated synthetic data.}
	\label{fig:R-S_comparison}
	\vspace{-4mm}
\end{figure*}

\vspace{-2mm}
\section{Results}
This section presents the achieved results by our proposed UsynthFace and its components, including: 1) Studying the effect of different data augmentation operations on identity preservation. 2) Analysing identity-shared information between synthetic data and original generative model authentic training data. 3) Ablation studies of different components of our framework (Figure \ref{fig:overview}), where we provide extensive experiment evaluations of the components of our framework (marked with red rectangles in Figure \ref{fig:overview}). 4) Comparison with SOTA synthetic-based FR.  

\subsection{To which degree does data augmentation effect identity information?}
We evaluated the effect of augmenting face images on the identity in the face image. The two augmented versions of the same image (instance) were considered as a genuine pair and pairing with any other image of different instances is considered as an imposter pair.  We used SOTA FR model ElasticFace (ElasticFace-Arc) \cite{ElasticFace} to extract representation features of our synthetic data.  The achieved verification performances are reported as Equal Error Rate (EER), FMR10, FMR100, and FMR1000, which are the lowest false non-match rate (FNMR) for a false match rate (FMR) $\leq$10.0\%, $\leq$1.0\% and $\leq$0.1\% respectively, along with plotting the genuine-imposter score distributions.
We made the following observation: 1) GAN-based augmentations (pose, illumination and expression) preserve to large degree the identity information of the augmented sample (0.0110 EER) as shown in Table \ref{tab:distribution_metrics} and Figure \ref{fig:DFG_distribution}. 2) Color and geometric transformations through RandAugment lead to degradation in verification performance (0.0967 EER) in comparison to GAN-based augmentation. 3) As expected, combining GAN-based with RandAugment achieve the lowest verification performance (0.1650 EER) in comparison to the GAN-based (0.0110 EER) and RandAugment (0.0967 EER). However, combining GAN-based with RandAugment (thus creating challenging genuine pairs) significantly improved our unsupervised model on the considered benchmarks as will be shown in Section \ref{sec:abl_RA_N_M}.

\begin{table}[h!]
\resizebox{\linewidth}{!}{%
\begin{tabular}{|c|c|c|c|c|c|}
\hline
\textbf{Augmentation} & \textbf{LFW} & \textbf{AgeDB-30} & \textbf{CFP-FP} & \textbf{CA-LFW} & \textbf{CP-LFW} \\ \hline
HF                    & 73.12        & 50.95             & 60.99           & 60.05           & 56.13           \\ \hline
HF+GAN-based          & \textbf{81.53}        & \textbf{53.65}             & \textbf{67.21}           & \textbf{65.03}           & \textbf{64.22}           \\ \hline
\end{tabular}%
}
\caption{Verification accuracies (\%) of two data augmentation settings on five different FR benchmarks. HF refers to horizontal-flipping. Adding GAN-based augmentations enhances the accuracy of the resulting FR model, which will be referred to as "baseline". Higher accuracy in bold.
%Results of model training with only horizontal-flipping and additional GAN-based augmentations.
}
\label{tab:HF_and_GAN}
\vspace{-4mm}
\end{table}

\subsection{Does the synthetic data share identity information with the GAN authentic training  data?}
Driven by privacy concerns, we answered this question by conducting an N:N evaluation where references were compared to probes from the GAN authentic training dataset (noted as R-R) and N:M evaluation where authentic references from the GAN training dataset were compared to synthetic probes generated by GAN generator model (noted as R-S). In this experiment, feature representations were obtained from ArcFace \cite{ArcFace}, CurricularFace \cite{CurricularFace} and ElasticFace \cite{ElasticFace}, respectively \footnote{The network architecture of ElasticFace-Arc \cite{ElasticFace}, ArcFace  \cite{ArcFace} and CurricularFace \cite{CurricularFace} is ResNet100 trained on MS1MV2 \cite{MS-Celeb-1M} by the corresponding authors (model publicly available).}. As authentic and synthetic datasets do not have identity label, we calculated the operation thresholds at FMR100 (FMR100\_Th) and FMR1000 (FMR1000\_Th) for each of the evaluated models on LFW dataset.
The comparison scores below the operational threshold were considered as non-match, i.e. of a different identity and the ones that were higher than the operational threshold were considered as match, i.e. of the same identity. 
Figure \ref{fig:R-S_comparison} shows score distributions of the cross-dataset (R-S) and authentic data (R-R) with two operational thresholds FMR100\_Th and  FMR1000\_Th. It can be clearly noticed that R-R and R-S score distributions are highly overlapped and only a few samples are considered matched, i.e. achieved comparison scores higher than the operational threshold. This observation is complementary to the previous findings in \cite{ThisFaceDoesNotExist} and \cite{SFace}, as these works also reported that the identity relation between the GAN training dataset and synthetic data is weak. 

\begin{figure}[h!]
	\centering
	\includegraphics[width=0.75\linewidth]{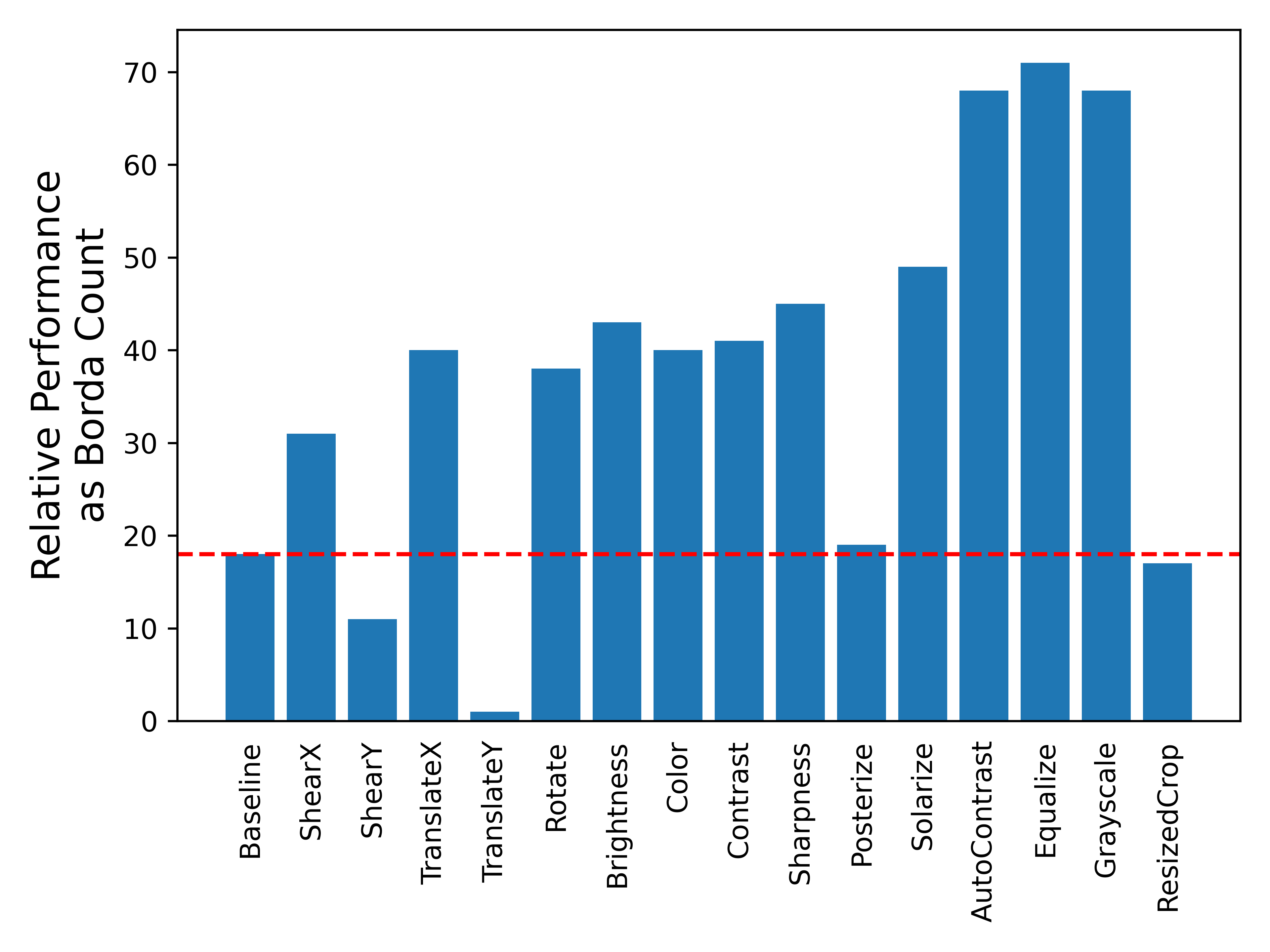}
	\vspace{-3mm}
    \caption{Augmentation operation selection for USynthFace indicated by the Borda Count (Table \ref{tab:RA_operations}) as verification performance. 12 out of 15 candidate operations outperformed baseline operations, composing the search space for RandAugment.}
	\label{fig:RA_operations}
	\vspace{-4mm}
\end{figure}

\subsection{Impact of GAN-based augmentation}
\label{sec:abl_impact_GAN_based}
We evaluated the impact of GAN-based augmentation on our USynthFace by training and evaluating USynthFace model with widely used augmentation operation in FR \cite{ElasticFace, ArcFace, CurricularFace, SphereFace}, horizontal-flipping. This model is considered as a baseline in this study. Then, we trained a second instance of the baseline model with GAN-based augmentation, i.e. pose, illumination and expression (in addition to horizontal-flipping). The achieved verification performances on the considered evaluation benchmarks are presented in Table \ref{tab:HF_and_GAN}. One can clearly notice that including GAN-based augmentation in the model training significantly improved the verification accuracies in comparison to the baseline model.

\begin{table}[h!]
\resizebox{\linewidth}{!}{%
\begin{tabular}{|l|c|c|c|c|c|c|}
\hline
\textbf{\begin{tabular}[c]{@{}l@{}}Augmentation\\ Operation\end{tabular}} & \textbf{LFW} & \textbf{AgeDB30} & \textbf{CFP-FP} & \textbf{CALFW} & \textbf{CPLFW} & \textbf{\begin{tabular}[c]{@{}c@{}}Borda \\ Count\end{tabular}} \\ \hline
Baseline                                                                  & 81.53        & 53.65            & 67.21           & 65.03          & 64.22          & 18                                                              \\ \hline \hline
ShearX                                                                    & 81.45        & 53.63            & 68.33           & 65.13          & 64.43          & 31                                                              \\ \hline
ShearY                                                                    & 80.63        & 53.12            & 67.57           & 63.45          & 64.28          & 11                                                              \\ \hline
TranslateX                                                                & 81.82        & 53.77            & 69.4            & 64.78          & 64.98          & 40                                                              \\ \hline
TranslateY                                                                & 76.55        & 52.77            & 67.39           & 61.03          & 62.73          & 1                                                               \\ \hline
Rotate                                                                    & 82.02        & 54.03            & 67.91           & 64.82          & 65.17          & 38                                                              \\ \hline
Brightness                                                                & 82.17        & 54.28            & 67.64           & 65.07          & 65.23          & 43                                                              \\ \hline
Color                                                                      & 81.73        & 56.12            & 67.71           & 66.05          & 64.93          & 40                                                              \\ \hline
Contrast                                                                  & 82.98        & 53.30            & 68.07           & 64.85          & 65.4           & 41                                                              \\ \hline
Sharpness                                                                 & 81.92        & 55.37            & 68.06           & 65.55          & 64.95          & 45                                                              \\ \hline
Posterize                                                                 & 80.47        & 54.12            & 67.61           & 64.48          & 64.33          & 19                                                              \\ \hline
Solarize                                                                  & 81.87        & 54.85            & 69.00           & 66.23          & 64.93          & 49                                                              \\ \hline
AutoContrast                                                              & 84.05        & 57.88            & 69.03           & 67.37          & 66.73          & 68                                                              \\ \hline
\textbf{Equalize}                                                                  & 85.23        & 58.57            & 69.63           & 67.45          & 65.42          & \textbf{71}                                                              \\ \hline
Grayscale                                                                 & 83.05        & 63.87            & 68.51           & 67.78          & 65.68          & 68                                                              \\ \hline
ResizedCrop                                                               & 78.92        & 53.02            & 68.33           & 63.35          & 64.37          & 17                                                              \\ \hline
\end{tabular}%
}
\caption{Impact of different conventional augmentation operations given as verification accuracies (\%) of the trained models. 
%The "baseline" model is the baseline. 
The borda count shows that many augmentations improve beyond the baseline model.}
\label{tab:RA_operations}
\vspace{-4mm}
\end{table}

\begin{figure}[h!]
	\centering
	\includegraphics[width=0.65\linewidth]{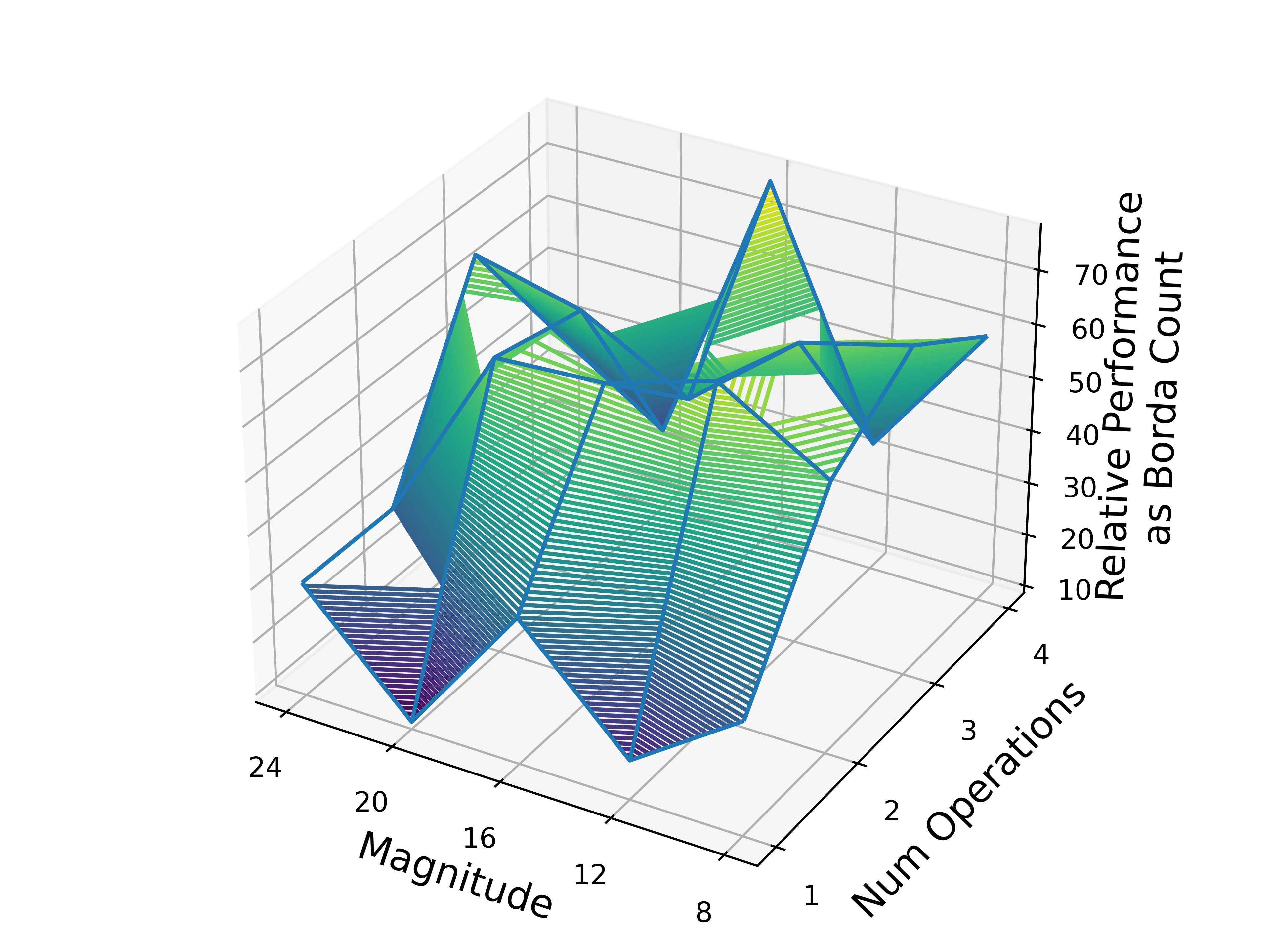}
    \caption{Hyperparameter of RandAugment selection for USynthFace in terms of Borda Count (Table \ref{tab:RA_hyperparameter}) as a verification performance. }
	\label{fig:RA_hyperparameters}
	\vspace{-5mm}
\end{figure}

\subsection{Impact of conventional data augmentation}
\label{sec:abl_RA_operations}
In this study, we evaluated the achieved verification performance by introducing different geometric/color transformations to model training.
The baseline model is noted as "Baseline" and trained with horizontal-flipping and GAN-based augmentations.
Table \ref{tab:RA_operations} and Figure \ref{fig:RA_operations} present the achieved results by different models, each was trained with a single candidate augmentation operation (in addition to horizontal-flipping and GAN-based operations). The candidate operation is included in the final augmentation space if it has led to improvement in overall verification performances (in terms of Borda count) in comparison to the baseline model. Out of 15 candidate operations, 12 operations outperformed the baseline operation. 
These operations are included in the search space of RandAugment.

\begin{table}[h!]
\resizebox{\linewidth}{!}{%
\begin{tabular}{|c|c|c|c|c|c|c|c|}
\hline
\multicolumn{1}{|l|}{\textbf{\begin{tabular}[c]{@{}l@{}}Number of\\ Operations\end{tabular}}} & \textbf{Magnitude} & \textbf{LFW} & \textbf{AgeDB30} & \textbf{CFP-FP} & \textbf{CALFW} & \textbf{CPLFW} & \textbf{\begin{tabular}[c]{@{}l@{}}Borda \\ Count\end{tabular}} \\ \hline
\multirow{5}{*}{1}                                                                            & 8                  & 86.43        & 63.48            & 72.97           & 68.83          & 68.02          & 30                                                              \\
                                                                                              & 12                 & 86.35        & 62.23            & 73.04           & 68.73          & 68.13          & 16                                                              \\
                                                                                              & 16                 & 86.50        & 65.13            & 73.39           & 68.37          & 68.37          & 36                                                              \\
                                                                                              & 20                 & 86.27        & 62.72            & 72.20           & 68.70          & 67.93          & 10                                                              \\
                                                                                              & 24                 & 86.48        & 62.80            & 73.64           & 68.72          & 68.63          & 30                                                              \\ \hline
\multirow{5}{*}{2}                                                                            & 8                  & 86.95        & 63.50            & 73.73           & 69.13          & 68.38          & 59                                                              \\
                                                                                              & 12                 & 87.00        & 63.33            & 74.31           & 69.43          & 68.75          & 71                                                              \\
                                                                                              & 16                 & 86.93        & 63.20            & 74.36           & 68.88          & 68.92          & 65                                                              \\
                                                                                              & 20                 & 86.98        & 62.98            & 74.31           & 69.17          & 68.77          & 64                                                              \\
                                                                                              & 24                 & 86.55        & 62.97            & 74.17           & 68.52          & 68.15          & 30                                                              \\ \hline
\multirow{5}{*}{3}                                                                            & 8                  & 86.92        & 63.95            & 74.01           & 69.67          & 68.88          & 70                                                              \\
                                                                                              & 12                 & 86.60        & 63.20            & 75.11           & 68.80          & 68.88          & 65                                                              \\
                                                                                              & 16                 & 86.68        & 63.22            & 74.21           & 68.75          & 68.42          & 49                                                              \\
                                                                                              & 20                 & 86.57        & 63.82            & 75.59           & 68.50          & 68.77          & 60                                                              \\
                                                                                              & 24                 & 86.57        & 63.13            & 74.54           & 68.88          & 69.15          & 65                                                              \\ \hline
\multirow{5}{*}{\textbf{4}}                                                                            & 8                  & 86.70        & 62.82            & 74.19           & 69.07          & 69.25          & 59                                                              \\
                                                                                              & 12                 & 86.53        & 62.67            & 74.37           & 68.47          & 68.63          & 33                                                              \\
                                                                                              & \textbf{16}                 & 86.93        & 64.15            & 74.51           & 69.08          & 68.80          & \textbf{77}                                                              \\
                                                                                              & 20                 & 86.25        & 62.72            & 74.57           & 68.08          & 68.02          & 24                                                              \\
                                                                                              & 24                 & 86.00        & 64.33            & 74.41           & 67.52          & 68.27          & 37                                                              \\ \hline
\end{tabular}%
}
\caption{Sensitivity study on RandAugmen hyperparameters. Verification accuracies (\%) of different settings on five FR benchmarks. Randomly applying 4 operations with the magnitude of 16 obtained the best overall performance. The highest Borda Count (accuracy) is in bold.}
\label{tab:RA_hyperparameter}
\vspace{-3mm}
\end{table}

\begin{table}[h!]
\resizebox{\linewidth}{!}{%
\begin{tabular}{|c|c|c|c|c|c|c|}
\hline
\textbf{\begin{tabular}[c]{@{}c@{}}Queue\\ Size\end{tabular}} & \textbf{LFW} & \textbf{AgeDB30} & \textbf{CFP-FP} & \textbf{CALFW} & \textbf{CPLFW} & \textbf{\begin{tabular}[c]{@{}c@{}}Borda \\ Count\end{tabular}} \\ \hline
512                                                           & 86.22        & 63.43            & 74.44           & 67.73          & 68.53          & 8                                                               \\ \hline
1024                                                          & 86.55        & 62.58            & 73.97           & 67.87          & 67.85          & 7                                                               \\ \hline
2048                                                          & 86.25        & 63.97            & 73.67           & 68.63          & 68.57          & 14                                                              \\ \hline
4096                                                          & 86.97        & 63.25            & 74.84           & 68.82          & 68.72          & 24                                                              \\ \hline
8192                                                          & 86.50        & 63.60            & 75.30           & 68.30          & 68.67          & 22                                                              \\ \hline
16384                                                         & 86.47        & 62.72            & 73.67           & 68.85          & 69.07          & 17                                                              \\ \hline
\textbf{32768}                                                & 86.93        & 64.15            & 74.51           & 69.08          & 68.80          & \textbf{30}                                                      \\ \hline
65536                                                         & 87.02        & 63.57            & 74.47           & 67.87          & 68.18          & 18                                                              \\ \hline
\end{tabular}%
}
\caption{Verification accuracies (\%) of using different queue sizes on five FR benchmarks. The highest accuracy indicated by the highest Borda Count is in bold. }
\label{tab:queue_size}
\vspace{-3mm}
\end{table}

\subsection{Conventional data augmentation through RandAugment}
\label{sec:abl_RA_N_M}
The augmentation operations from the previous study (Section \ref{sec:abl_RA_operations}) are used to build the search space for RandAugment. We evaluate in this section by randomly augmenting the training samples with multiple operations, i.e. 1, 2, 3 or 4 and with different magnitudes, i.e. 8, 12, 16, 20 or 24. In total, we trained and evaluated 20 models (4 different numbers of operations and five possible magnitudes). The best verification performance is achieved by randomly applying 4 operations (sequentially) with the magnitude of 16 as shown in Table \ref{tab:RA_hyperparameter} and Figure \ref{fig:RA_hyperparameters}.

\begin{table}[h!]
\resizebox{\linewidth}{!}{%
\begin{tabular}{|c|c|c|c|c|c|c|}
\hline
\textbf{\begin{tabular}[c]{@{}c@{}}Feature\\ Dimensionality\end{tabular}} & \textbf{LFW} & \textbf{AgeDB-30} & \textbf{CFP-FP} & \textbf{CA-LFW} & \textbf{CP-LFW} & \textbf{\begin{tabular}[c]{@{}c@{}}Borda\\ Count\end{tabular}} \\ \hline
128                                                                       & 86.35        & 63.55             & 73.93           & 68.52           & 68.12           & 9                                                              \\ \hline
256                                                                       & 86.52        & 63.02             & 74.23           & 68.48           & 68.33           & 10                                                             \\ \hline
\textbf{512}                                                             & 86.93        & 64.15             & 74.51           & 69.08           & 68.80           & \textbf{20}                                                     \\ \hline
1024                                                                      & 86.65        & 63.52             & 73.71           & 68.75           & 68.15           & 11                                                             \\ \hline
\end{tabular}%
}
\caption{Verification accuracies (\%) achieved by models with different feature representation dimensionality on five FR benchmarks. The best overall verification accuracy is achieved by feature representation of 512-D.}
\label{tab:emb_size}
\vspace{-4mm}
\end{table}

\begin{table}[h!]
\resizebox{\linewidth}{!}{%
\begin{tabular}{|c|c|c|c|c|c|c|}
\hline
\textbf{\begin{tabular}[c]{@{}c@{}}LR-\\ Scheduler\end{tabular}} & \textbf{\begin{tabular}[c]{@{}c@{}}Maximal\\ Epochs\end{tabular}} & \textbf{LFW} & \textbf{AgeDB-30} & \textbf{CFP-FP} & \textbf{CA-LFW} & \textbf{CP-LFW}  \\ \hline
step-based                                                       & 40                                                                & 86.93        & 64.15             & 74.51           & 69.08           & 68.80                                                                       \\ \hline
plateau-based                                                    & 200                                                               & \textbf{91.52}        & \textbf{69.30  }           & \textbf{78.46}           & \textbf{75.35 }          & \textbf{71.93 }                                                             \\ \hline
\end{tabular}%
}
\caption{Verification accuracies (\%) of different LR schedulers on five FR benchmarks. Models trained with plateau-based LR scheduler and more epochs yield better performances than step-based scheduler.}
\label{tab:training_optim}
\vspace{-3mm}
\end{table}

\begin{table}[h!]
\resizebox{\linewidth}{!}{%
\begin{tabular}{|c|c|c|c|c|c|c|}
\hline
\textbf{Margin} & \textbf{LFW} & \textbf{AgeDB-30} & \textbf{CFP-FP} & \textbf{CA-LFW} & \textbf{CP-LFW} & \textbf{\begin{tabular}[c]{@{}c@{}}Borda\\ Count\end{tabular}} \\ \hline
0.00            & 91.52        & 69.30             & 78.46           & 75.35           & 71.93           & 12                                                             \\ \hline
0.05            & 91.30        & 70.37             & 78.73           & 75.52           & 71.58           & 13                                                             \\ \hline
\textbf{0.10}   & 92.12        & 71.08             & 78.19           & 76.15           & 71.95           & \textbf{22}                                                             \\ \hline
0.15            & 91.83        & 70.78             & 78.11           & 76.18           & 71.50           & 17                                                             \\ \hline
0.20            & 91.65        & 70.75             & 77.80           & 75.93           & 71.37           & 11                                                             \\ \hline
\end{tabular}%
}
\caption{Verification accuracies (\%) of different margin values for the MarginNCE loss on five FR benchmarks. A margin value of 0.10 leads to the best overall performance and thus is used in the final experimental setting.}
\label{tab:margin}
\vspace{-4mm}
\end{table}

\begin{table*}[h!]
\centering
\begin{tabular}{|c|c|c|c|c|c|c|c|c|c|}
\hline
\textbf{Method}               & \textbf{Unsupervised} & \textbf{Identities} & \textbf{\begin{tabular}[c]{@{}c@{}}Samples per\\ Identity\end{tabular}} & \textbf{Total} & \textbf{LFW}   & \textbf{AgeDB-30} & \textbf{CFP-FP} & \textbf{CA-LFW} & \textbf{CP-LFW} \\ \hline
SynFace \cite{SynFace}        & \FeatureFalse         & 10K                 & 50                                                                      & 500K           & 88.98          & -                 & -               & -               & -               \\
SynFace (w/IM) \cite{SynFace} & \FeatureFalse         & 10K                 & 50                                                                      & 500K           & 91.97          & -                 & -               & -               & -               \\ \hline
SFace-10 \cite{SFace}         & \FeatureFalse         & 10,575              & 10                                                                      & 105K           & 87.13          & 63.30             & 68.84           & 73.47           & 66.82           \\
SFace-20 \cite{SFace}         & \FeatureFalse         & 10,575              & 20                                                                      & 211K           & 90.50          & 69.17             & 73.33           & 76.35           & 71.17           \\
SFace-40 \cite{SFace}         & \FeatureFalse         & 10,575              & 40                                                                      & 423K           & 91.43          & 69.87             & 73.10           & 76.92           & \textbf{73.42}  \\
SFace-60 \cite{SFace}         & \FeatureFalse         & 10,575              & 60                                                                      & 634K           & 91.87          & \textbf{71.68}    & 73.86           & \textbf{77.93}  & 73.20           \\ \hline
USynthFace (ours)              & \FeatureTrue          & 100K                & 1                                                                       & 100K           & 91.52          & 69.30             & 78.46           & 75.35           & 71.93           \\
USynthFace (ours)              & \FeatureTrue          & 200K                & 1                                                                       & 200K           & 91.93          & 71.23             & 78.03           & 76.73           & 72.27           \\
USynthFace (ours)              & \FeatureTrue          & 400K                & 1                                                                       & 400K           & \textbf{92.23} & 71.62             & \textbf{78.56}  & 77.05           & 72.03           \\ \hline
\end{tabular}
\caption{Verification accuracies (\%) on five different FR benchmarks achieved by the supervised and SOTA SynFace \cite{SynFace} and SFace \cite{SFace} models, and our USynthFace model trained on the synthetic training databases of different sizes. The bold number refers to the highest performance on each benchmark. Nothing that the authors of SynFace \cite{SynFace} only provided evaluation results on LFW. Our unsupervised USynthFace model obtained very competitive and even better results than supervised synthetic-based FR models. }
\label{tab:comparison_SOTA}
\vspace{-4mm}
\end{table*}

\subsection{Analyses of the queue size }
\label{sec:abl_queue_size}
In the previous section, we evaluated several augmentation methods for training our USynthFace model. In this section, we study varying the queue size of momentum contrast. Our achieved results in Table \ref{tab:queue_size} pointed out that maintaining a queue of 32768 negative keys leads to the highest overall verification performance on the considered evaluation benchmarks. Noting that increasing the queue size to 65536 did not improve the overall verification performances.

%\begin{figure}[ht]
%	\centering
%	\includegraphics[width=0.6\linewidth]{figures/disco_queue_size.png}
 %   \caption{The queue size selection for the proposed USynthFace in terms of Borda Count (Table \ref{tab:queue_size}) as a verification performance. The queue size of 32768 yields the best performance.}
%	\label{fig:queue_size}
%	\vspace{-4mm}
%\end{figure}

\subsection{Study of feature representation dimensionality}
\label{sec:abl_feat_dim}
Based on the optimal queue size and augmentation methods, we evaluate in this section different learned feature representation dimensionalities i.e. 128, 256, 512 and 1024. All models in this section are trained with GAN-based augmentation and RandAugment with 4 sequential operations and a magnitude of 16 as well as a queue size of 32668. The achieved results of utilizing different feature representation dimensionality are presented in Table \ref{tab:emb_size} where the best overall verification performance is achieved using feature representation of 512-D.

\subsection{Training optimization}
\label{sec:abl_training_optim}
The presented results are achieved so far by training our USynthFace models using a step-based learning rate schedule. Previous works \cite{MocoV2} on unsupervised representation learning pointed out that increasing the number of training epochs is beneficial for improving unsupervised model accuracy.  
To provide complete evaluation results, we study increasing the number of epochs to a maximum of 200 \cite{MocoV2} and using a plateau-based learning scheduler. The  initial learning rate is set to 0.1 and it divided by 10 when the average validation accuracy does not improve for 10 consecutive epochs. The training is stopped when the average validation accuracy does not improve for 20 consecutive epochs with maximum of 200 epochs. Using the listed training settings, the model training stopped after 91 training epochs. The achieved results, in this case, are presented in Table \ref{tab:training_optim}, pointing out that increasing the training epochs significantly improved the verification performance on all considered benchmarks.

%We noticed that some of the models do not fully converge in the given training time. Hence, we change the learning rate scheduler from a step-based scheduler to a plateau-based scheduler. The learning rate is divided by 10 when the average validation accuracy on LFW, AgeDB-30 and CFP-FP does not improve for 10 epochs. The training is stopped when the same average validation accuracy does not improve for 20 epochs. The maximum number of epochs is set to 200, but it was never reached in the presented experiments.

\subsection{Impact of different margins in MarginNCE}
\label{sec:abl_margin}
We study in this section different margin values (0, 0.05, 0.1, 0.15 and 0.2) for MarginNCE loss (Eq. \ref{eq:MarginNCE}). The presented results in this section were obtained by training four different models with the optimal observed training settings from the previous experiments. It can be noticed from the achieved results in Table \ref{tab:margin} that the overall verification performance is improved by increasing margin values from 0 to 0.1. However, when we increase the margin value to 0.15 or 0.20, the overall verification performances are slightly degraded.

\subsection{Study of training database size}
\label{sec:abl_db_size}
Given that there is no restriction on the number of synthetic samples that can be generated using the generative model, we increased the training dataset size from 100K to 200K and to 400K images. Then, we trained two instances of our unsupervised model with the new constructed datasets. The achieved results are presented in Table \ref{tab:comparison_SOTA} together with other SOTA synthetic-based FR models. One can notice that increasing the training dataset sizes to 200K and 400K images slightly improved the overall verification performance in comparison to the model trained with 100K images. 

%\begin{table}[ht]
%\resizebox{\linewidth}{!}{%
%\begin{tabular}{|c|c|c|c|c|c|c|}
%\hline
%\textbf{\begin{tabular}[c]{@{}c@{}}Database\\ Size\end{tabular}} & \textbf{LFW} & \textbf{AgeDB-30} & \textbf{CFP-FP} & \textbf{CA-LFW} & \textbf{CP-LFW} & \textbf{\begin{tabular}[c]{@{}c@{}}Borda\\ Count\end{tabular}} \\ \hline
%100K                                                             & 92.12        & 71.08             & 78.19           & 76.15           & 71.95           & 7                                                             \\ \hline
%200K                                                             & 91.93        & 71.23             & 78.03           & 76.73           & 72.27           & 8                                                             \\ \hline
%\textbf{400K}                                                    & 92.23        & 71.62             & 78.56           & 77.05           & 72.03           & \textbf{14}                                                    \\ \hline
%\end{tabular}%
%}
%\caption{Study of different database sizes on GAN-based and RandAugment augmentations}
%\label{tab:db_size}
%\end{table}

\subsection{Comparison with the SOTA synthetic-based FR}
We compared the achieved verification performance by our USynthFace with the recent SOTA FR that proposed the use of synthetic data in FR training (Table \ref{tab:comparison_SOTA}). Noting that this is the first work that proposed to train FR with privacy-friendly synthetic data in an unsupervised fashion. SynFace \cite{SynFace} and SFace \cite{SFace} are trained with supervised learning to learn multi-class classification using margin-penalty softmax loss. SynFace only reported the verification performance on LFW dataset.
On LFW and CFP-FP datasets, our unsupervised model outperformed SFace and SynFace. On AgeDB-30, CA-LFW and CP-LFW, our unsupervised model achieved very competitive results to SFace, even though USynthFace training is unsupervised.

\section{Conclusion}
We presented in this work a novel unsupervised face recognition solution trained on unlabeled synthetic data.
%based on the concept of Momentum Contrast and contrastive learning.
The unsupervised training is based on creating positive pairs to unlabeled synthetic face images of random identities through well-studied augmentations.
We proposed not only to use conventional data augmentations in our USynthFace model training, but also introduced GAN-based augmentation to the training pipeline, enhancing the variability in the synthetic face image appearances.
This has been complemented with a set of empirical studies on the validity of the different components of our USynthFace and their design choices. With a simple yet effective training paradigm, our USynthFace advanced the SOTA performance on a number of the evaluation benchmarks, in comparison to the recent face recognition models trained on synthetic data, while being the only one trained in an unsupervised manner.

\vspace{-2mm}

{\small
\bibliographystyle{ieee}
\bibliography{egbib}
}

\end{document}